\documentclass[final]{arxiv}

\usepackage{amssymb}
\usepackage{amsmath}
\usepackage{graphicx}
\usepackage{float}
\usepackage{algorithm, algorithmic}
\usepackage{color}
\usepackage{widetext}
\usepackage{subcaption}
\usepackage{pbox}
\usepackage[pdfencoding=auto]{hyperref}
\usepackage{bookmark}
\bookmarksetup{numbered,open}

\begin{document}

\title{A Neurodynamical System for finding a Minimal VC Dimension Classifier}

\author{Jayadeva$^a$, Sumit Soman$^a$, Amit Bhaya$^b$ \\
$^a$ Department of Electrical Engineering, Indian Institute of Technology, Delhi, India \\
$^b$ Department of Electrical Engineering (PEE), Graduate School of Engineering (COPPE), Federal University of Rio de Janeiro (UFRJ), Rio de Janeiro, Brazil \\
\textit{E-mail: jayadeva@ee.iitd.ac.in, sumit.soman@gmail.com, amit@nacad.ufrj.br}
}

\begin{abstract}
The recently proposed Minimal Complexity Machine (MCM) finds a hyperplane classifier by minimizing an exact bound on the Vapnik-Chervonenkis (VC) dimension. The VC dimension measures the capacity of a learning machine, and a smaller VC dimension leads to improved generalization. On many benchmark datasets, the MCM generalizes better than SVMs and uses far fewer support vectors than the number used by SVMs. In this paper, we describe a neural network based on a linear dynamical system, that converges to the MCM solution. The proposed MCM dynamical system is conducive to an analogue circuit implementation on a chip or simulation using Ordinary Differential Equation (ODE) solvers. Numerical experiments on benchmark datasets from the UCI repository show that the proposed approach is scalable and accurate, as we obtain improved accuracies and fewer number of support vectors (upto 74.3\% reduction) with the MCM dynamical system.
\end{abstract}

\maketitle


\section{Introduction\label{sec:intro}}

Support vector machines (SVMs) have evolved to become one of the most widely used machine learning techniques today owing. They have also been employed for a number of applications to obtain cutting edge performance; novel uses have also been devised, where their utility has been amply demonstrated. The classical SVM \cite{L1svm} and the least squares SVM (LSSVM) \cite{suykens1999least} have spawned a multitude of  formulations.  Most SVM formulations require the solution of a Quadratic Programming Problem (QPP), involving an objective function maximizing the margin (with a term for the admissible error in case of soft-margin SVM) and suitable constraints. The solution to such an optimization problem is obtained in terms of a separating hyperplane, the determination of which is a direct consequence of the number of support vectors identified in the dataset. Practical machine learning problems of today involve large datasets, and efficient real-time performance of learning systems demands the use of learning algorithms which minimize the learning complexity, in terms of space, time or both.

The complexity of learning systems, such as SVMs, can be estimated by the Vapnik-Chervonenkis (VC) dimension. A smaller value of the VC dimension indicates robust generalization and lower test set error rates; hence a large VC dimension would be undesirable. As stated in pioneering work by Vapnik \cite{vapnik98}, Burges \cite{burges1998} and others, SVMs can have a large, possibly infinite VC dimension, which could also be infinite. This implies that SVMs may work well in practice, but there is no guarantee that they will generalize well. In fact, Vapnik and Chervonenkis \cite{vapnik1974theory} arrive at a bound on the stochastic approximation of the empirical risk, as given by Equations (\ref{eqn:vc_dim_1})-(\ref{eqn:vc_dim_2}), which holds with probability $(1-\eta)$.

\begin{eqnarray}
R(\lambda) \leq R_{emp}(\lambda) + \sqrt{\frac{h (ln \frac{2l}{h} + 1) - ln\frac{\eta}{4}}{l}} \label{eqn:vc_dim_1}\\
\text{Where, } \; R_{emp}(\lambda) = \frac{1}{l}	\sum_{i=1}^l |f_\lambda(x_i)-y_i|, \label{eqn:vc_dim_2}
\end{eqnarray}

\noindent and $f_\lambda$ is a function having VC-dimension $h$ with the smallest empirical risk on a dataset $\lbrace x_i, i= 1,2,...,l\rbrace$ of $l$ data points with corresponding labels $\lbrace y_i, i= 1,2,...,l\rbrace$.

Recently, it has been shown that a formulation termed as the Minimal Complexity Machine (MCM)  \cite{jd2014b} can be used to realize a large-margin classifier while minimizing an exact (\boldmath{$\Theta$}) bound on the VC dimension. The approach requires the solution of a linear programming problem, and generalizes well on benchmark datasets. The MCM outperforms SVMs in terms of test set accuracy, while using far fewer support vectors; in many instances, the MCM predicts better while using less than 10\% the number of support vectors used by SVMs \cite[Table III]{jd2014b}. Variants of the MCM have been proposed for regression \cite{mcm_regress}, fuzzy classification \cite{mcm_fuzzy} and feature selection for large datasets \cite{mcm_featsel}.

Our focus in this paper is a neurodynamical system that converges to the MCM solution, thus yielding a minimal VC dimension classifier. A dynamical system that converges to a minimum VC dimension classifier allows for high speed and real-time implementation, e.g. as an analogue VLSI chip. Since this approach yields a system that has low complexity, it opens a large vista of applications in the learning and modelling domains. The MCM solutions are usually very sparse; this provides the advantage of lower computational cost in a hardware implementation. These advantages carry over to VLSI implementations and are therefore of much interest.

Applications based on dynamical systems have attracted significant attention over the last three decades, owing to the potential for real time, high speed realizations as electronic circuits \cite{rodriguez1990nonlinear} or as recurrent neural networks \cite{wang1993analysis, xia1998general}. The behaviour of such neurodynamical systems is also interesting as it has been used in modeling biological systems \cite{freeman2007indirect, marupaka2012connectivity, corchs2001neurodynamical}, solving optimization problems \cite{yan2014collective, tank1986simple, chua1984nonlinear, brockett1988dynamical}, large-scale problems \cite{hasegawa2002solving}, fuzzy symbolic dynamics \cite{dobosz2010understanding} and working memory \cite{pascanu2011neurodynamical} among others.

There have also been several works which integrate the linear/quadratic programming approach within Neurodynamical systems. For instance, Bennett and Mangasarian \cite{mangasarian1992neural} proposed a technique for training neural networks using linear programming based on the Multi-surface Method, which was applied for breast cancer diagnosis. Faybusovich \cite{faybusovich1992dynamical, faybusovich1991dynamical, faybusovich1991hamiltonian} proposed dynamical systems for solving linear programming based on barrier functions and presented their Hamiltonian analysis. Maa and Shanblatt \cite{maa1992linear} present a neural network formulation for linear and quadratic programming, extending the network originally proposed by Kennedy and Chua \cite{kennedy1988neural}. Jun Wang presented a recurrent neural network for solving Linear Programming Problems (LPP) \cite{wang1993analysis} in 1993, which was followed by a neural network for solving LPPs with bounded variables by  Xia and Wang \cite{xia1995neural} in 1995. In 1996, Wu et al. presented a neural network with global convergence guarantees \cite{wu1996high, xia1996new}. Other work in this direction includes the approaches presented by Oskoei and Amiri in 2006 \cite{ghasabi2006efficient} and by Chukwunenye in 2014 \cite{chukwunenyeinterior}. An overview of dynamical system methods for mathematical programming from a control perspective can be found in Bhaya and Kaszkurewicz \cite{bhaya2006control}.

Recent work on the application of dynamical systems involves solving LPs for estimation in the context of image restoration by Xia et al.\cite{xia2012discrete} and solving the assignment problem \cite{hu2012solving}. Liu et al. \cite{liu2013one} demonstrate the use of a neural network to solve a non-smooth optimization problem with linear constraints, while P{\'e}rez-Ilzarbe \cite{perez2013new} shows its use for solving a quadratic problem with linear constraints. In contrast the MCM formulation allows us to find a minimal VC dimension classifier utilizing a neurodynamical system that finds the optimal solution of a LPP, with guaranteed convergence and provably good generalization.

The rest of the paper is organized as follows. Section \ref{sec2} introduces the Minimal Complexity Machine (MCM) and the associated optimization problem. Section \ref{sec3} describes the MCM neurodynamical system, and an analysis of its convergence on synthetic datasets is shown in Section \ref{sec4}. Section \ref{results} discusses simulation results. Section \ref{conclusion} contains concluding remarks.

\section{Motivating the Minimal Complexity Machine} \label{sec2}
Consider such a binary classification problem with data points $x^i, i = 1, 2, ..., M$, and where samples of class +1 and -1 are associated with labels $y_i = 1$ and $y_i = -1$, respectively. We assume that the dimension of the input samples is $n$, i.e. $x^i = (x_1^i, x_2^i, ..., x_n^i)^T$. The problem of interest is finding a hyperplane of the form
\begin{equation}
 u^Tx + v = 0.
\end{equation}
that has the smallest Vapnik-Chervonenkis dimension $\gamma$, and that separates the samples with least error. In \cite{ jd2014b}, it has been shown that there exist constants $\alpha, \beta > 0$, $\alpha, \beta \in \mathbb{R}$ such that
\begin{equation}\label{exactbound}
 \alpha h^2 \leq \gamma \leq \beta h^2,
\end{equation}
where
\begin{gather}
 h = \frac{\operatorname*{Max}_{i = 1, 2, ..., M} \|u^T x^i + v\|}{\operatorname*{Min}_{i = 1, 2, ..., M} \|u^T x^i + v\|}.
\end{gather}
In other words, $h^2$ constitutes a tight or exact ($\theta$) bound on the VC dimension $\gamma$. An exact bound implies that $h^2$ and $\gamma$ are close to each other. Thus, the machine capacity can be minimized by minimizing $h^2$, or equivalently, $h$. The MCM optimization problem attempts to find a classifier with the smallest machine capacity, that makes as few misclassification errors on the training data as possible. This leads to a fractional programming problem, which, after suitable transformations, leads to the following optimization problem \cite{ jd2014b}. This transformation is discussed in detail in \cite[App. A]{jd2014b}.

\begin{gather}
\operatorname*{Min}_{w, b, h} ~~h + C \cdot \sum_{i = 1}^M q_i \label{obj5}\\
h \geq y_i \cdot [{w^T x^i + b}] + q_i, ~i = 1, 2, ..., M \label{cons51}\\
y_i \cdot [{w^T x^i + b}] + q_i \geq 1, ~i = 1, 2, ..., M \label{cons52} \\
q_i \geq 0, ~i = 1, 2, ..., M. \label{cons53}
\end{gather}
Here, the choice of $C$ allows a tradeoff between the complexity (machine capacity) of the classifier and the classification error. The soft margin MCM is described by the formulation Equations (\ref{obj5})-(\ref{cons53}).

Once $w$ and $b$ have been determined by solving Equations (\ref{obj5})-(\ref{cons53}), the class of a test sample $x$ may be determined as before by using the sign of $f(x)$ in Equation (\ref{testresult}). 

\begin{equation}\label{testresult}
 f(x) = w^T x + b
\end{equation}
In (\ref{sec3}), we show how the MCM solution can be determined by a dynamical system.

On similar lines, the kernel MCM obtains a hyperplane in $\phi$ space given by
\begin{equation}
 f(x) = w^T \phi(x) + b
\end{equation}
where $\phi()$ maps input vectors into a higher dimensional image space. The kernel MCM solves the following optimization problem.

\begin{gather}
\operatorname*{Min}_{w, b, h, q} \; h + C \cdot \sum_{i = 1}^M q_i \label{objk7}\\
h \geq y_i \cdot \left[\sum_{j = 1}^M \lambda_j K(x^i, x^j) + b\right] + q_i, ~i = 1, 2, ..., M\\
y_i \cdot \left[\sum_{j = 1}^M \lambda_j K(x^i, x^j) + b\right] + q_i \geq 1, ~i = 1, 2, ..., M \\ \label{consk71}
q_i \geq 0, ~i = 1, 2, ..., M.
\end{gather}

Once the variables $\lambda_j, j = 1, 2, ..., M$ and $b$ are obtained, the class that a test point $x$ belongs to can be determined by evaluating the sign of
\begin{equation}
f(x) ~=~ w^T \phi(x) + b ~=~ \sum_{j = 1}^M \lambda_j K(x, x^j) + b.
\end{equation}\label{testval}

\section{The MCM neurodynamical system} \label{sec3}

The MCM implementation follows the approach of Nguyen \cite{nguyan2000nonlinear}, which solves a simple system of differential equations involving both primal and dual variables. 
Consider a linear programming problem in the standard form as given by Equations (\ref{lp01})-(\ref{lp03}).

\begin{gather}
\max_{\theta} \; q^{T} \theta \label{lp01} \\
\mbox{s.t.}~G\theta \leq p \label{lp02} \\
\mbox{and}~\theta \geq 0 \label{lp03}
\end{gather}

The dual is given by Equations (\ref{lp04}) - (\ref{lp06}).
\begin{gather}
\min_{\delta} \; p^{T} \delta \label{lp04} \\
\mbox{s.t.}~ G^T \delta \geq q \label{lp05} \\
\mbox{and}~\delta \geq 0 \label{lp06}
\end{gather}

where $\theta,q \in \mathbb{R}^{n}$, $G \in \mathbb{R}^{m \times n}$ and $\delta, p \in \mathbb{R}^{m}$.

The primal (resp.~dual) network variables  are denoted $\theta$(resp.~$\delta$) and evolve in time as described by the  pair of coupled linear ODEs in Equations (\ref{df01})-(\ref{df02}).
\begin{gather}
\frac{d\theta}{dt} = p - G^{T} (\delta + k \frac{d\delta}{dt}) \label{df01} \\
\frac{d\delta}{dt} = -q - G (\theta + k \frac{d\theta}{dt}) \label{df02}
\end{gather}
where $k$ is a positive constant.

Supposing for the moment that the neural network defined by the Equations (\ref{df01})-(\ref{df02}) converges to an equilibrium, it can be shown that the optimal solution for the primal and dual formulations in Equations (\ref{lp01})-(\ref{lp06}) is an equilibrium of (\ref{df01})-(\ref{df02}), as follows:

Let the $i^{th}$ element of $\theta$ be denoted as $\theta_{i}$. The Equation (\ref{df01}) can be written as 

\[
 \frac{d\theta_{i}}{dt} =
  \begin{cases}
   (p - G^{T} (\delta + k \frac{d\delta}{dt}))_{i}, & \text{if } \theta_{i} > 0 \; \forall i \\
   \max \lbrace (p - G^{T} (\delta + k \frac{d\delta}{dt}))_{i}, 0 \rbrace, & \text{if } \theta_{i} = 0 \; \forall i 
  \end{cases}
\]

If the equilibrium solution is represented as $\theta^{*}$ and $\delta^{*}$, then $\frac{d\theta
^{*}}{dt} = 0$ and $\frac{d\delta^{*}}{dt} = 0$. Thus, for all $i$,

\begin{equation}
(p - G^{T}\delta^{*})_{i} = 0, \; \text{if} \; \theta^{*}_{i} \geq 0 \label{eqn01}
\end{equation}
and,
\begin{equation}
(p - G^{T}\delta^{*})_{i} \leq 0, \; \text{if} \; \theta^{*}_{i} = 0 \label{eqn02}
\end{equation}
Further, for all $i$,
\begin{equation}
p - G^{T}\delta^{*} \leq 0 \label{eqn03}
\end{equation}
and,
\begin{equation}
G\theta^{*} - q \leq 0
\end{equation}

Hence $\theta^{*}$ and $\delta^{*}$ are feasible solutions for the system defined by Equations (\ref{df01})-(\ref{df02}). Also, we have
\begin{equation}
p^{T} \theta^{*} - \theta^{*} G^{T} \delta^{*} = 0 \label{eqn04}
\end{equation}
and,
\begin{equation}
\theta^{*} G^{T} \delta^{*} - q^{T} \delta^{*} = 0 \label{eqn05}
\end{equation}
which implies
\begin{equation}
p^{T} \theta^{*} = q^{T} \delta^{*}
\end{equation}

Hence $\theta^{*}$ and $\delta^{*}$ are optimal solutions for the system defined by Equations (\ref{df01})-(\ref{df02}).

Also, differentiating Equations (\ref{df01})-(\ref{df02}) we can write
\begin{eqnarray}
\ddot{\theta} & = -{G}^T (\dot{\delta} + k \ddot{\delta}) \label{thetadd}\\
\ddot{\delta} & = -{G} (\dot{\theta} + k \ddot{\theta}) \label{deltadd}
\end{eqnarray}

It remains to be proved that convergence to the equilibrium occurs. Eliminating $\delta$ (resp. $\theta$) from Equations (\ref{thetadd})-(\ref{deltadd}) yields a second order differential equation in $\theta$ (resp.$\delta$), namely:
\begin{align}
	(k^2 {G}^T {G} - {I}) \ddot{\theta} + 2k {G}^T {G}\dot{\theta} + {G}^T {G} {\theta} & = -{G}^T {q} \label{theta2ode}\\
	(k^2 {G} {G}^T - {I}) \ddot{\delta} + 2k {G} {G}^T\dot{\delta} + {G} {G}^T {\delta} & = -{G} {p} \label{delta2ode}
\end{align}

The asymptotic stability of these second order linear ODEs is determined by the properties of the coefficient matrices. For example, using \cite[Thm.~1]{bernstein1995lyapunov}, it follows that if $k$ is chosen large enough to make the matrix $k^2 {G}^T {G} - {I}$ positive definite and if ${G}^T {G}$ is positive definite, then (Equation (\ref{theta2ode})) is asymptotically stable, implying that, from all initial conditions, its trajectories converge to the equilibrium point $\theta^*$ (see Equation (\ref{eqn01}) ff.). One may note here that the assumption of one of the matrices $G^TG$ or $GG^T$ being positive definite is a mild one, since it corresponds to assuming that there are no redundant inequality constraints. Finally if $\theta$ converges, so must $\delta$, since we are assuming that both the primal and dual problems are feasible.

Hence, for the MCM, the system of equations that finds a minimum VC dimension classifier aims at finding the equilibrium solution for the set of variables represented by the augmented vector $X=[w, b, q, h]$.  As mentioned initially, we consider data points $x^i, i = 1, 2, ..., M$, associated with labels $y^i \in \lbrace +1,-1 \rbrace, i=1,2, ..., M$, and each data-point being $n$-dimensional. Let the set of data points be denoted by $\psi_{M \times n}$, of which each row corresponds to $x^i$, and the label vector be denoted by a diagonal matrix $\Upsilon$, with diagonal entries $y_i$s, i.e. $\Upsilon = diag(y_1, y_2, y_3, ..., y_M)$.

The system finds a solution for each of the $(M+n+2)$ variables, as shown in Equation (\ref{eqn:mcm_nn_1}). Further, the LPP, as shown by Equations (\ref{lp01})-(\ref{lp03}), will now be defined by $q$, $G$ and $p$ as denoted by Equations (\ref{eqn:mcm_nn_2})-(\ref{eqn:mcm_nn_4}), where the notation $[\Upsilon \cdot \psi]$ represents the multiplication of matrices $\Upsilon$ and $\psi$; and $I$ represents the identity matrix. 

\begin{figure*}
\begin{gather}
X=\begin{pmatrix}
  w_{(1 \times n)} & b_{(1 \times 1)} & q_{(1 \times M)} & h_{(1 \times 1)}
 \end{pmatrix}^T \label{eqn:mcm_nn_1}\\
q=\begin{pmatrix}
[0]_{(1 \times n)} & [0]_{(1 \times 1)} & C\times[1]_{(1 \times M)} & 1_{(1 \times 1)}
 \end{pmatrix}^T \label{eqn:mcm_nn_2}\\ 
G=\begin{pmatrix}
[\Upsilon \cdot \psi]_{(M \times n)} & y_{(M \times 1)} & [0]_{(M \times M)} & -[1]_{(M \times 1)} \\
-[\Upsilon \cdot \psi]_{(M \times n)} & -y_{(M \times 1)} & -[I]_{(M \times M)} & [0]_{(M \times 1)}
\end{pmatrix} \label{eqn:mcm_nn_3} \\
p=\begin{pmatrix}
[0]_{(M \times 1)} & -[1]_{(M \times 1)}
\end{pmatrix}^T \label{eqn:mcm_nn_4}
\end{gather}
\end{figure*}

The system to be solved can now be represented as shown by Equations (\ref{mcmdf01})-(\ref{mcmdf02}), where $k \geq 0$ is a free parameter that can be tuned, and $Z$ is the dual variable of $X$.

\begin{gather}
\frac{dX}{dt} = p - G^{T} (Z + k \frac{dZ}{dt}) \label{mcmdf01} \\
\frac{dZ}{dt} = -q - G (X + k \frac{dX}{dt}) \label{mcmdf02}
\end{gather}

For the kernel MCM, the formulation can be obtained similarly by using $\phi(x^i)$, where $\phi()$ is a mapping to the chosen kernel space. The matrices $X$, $q$, $G$ and $p$ are then represented as shown in Equations (\ref{eqn:kmcm_nn_2})-(\ref{eqn:kmcm_nn_4}).

\begin{figure*}
\begin{gather}
X=\begin{pmatrix}
  w_{(1 \times M)} & b_{(1 \times 1)} & q_{(1 \times M)} & h_{(1 \times 1)}
 \end{pmatrix}^T \label{eqn:kmcm_nn_1}\\
q=\begin{pmatrix}
[0]_{(1 \times M)} & [0]_{(1 \times 1)} & C\times[1]_{(1 \times M)} & 1_{(1 \times 1)}
 \end{pmatrix}^T \label{eqn:kmcm_nn_2}\\ 
G=\begin{pmatrix}
[\Upsilon [\phi(x^i)^T \phi(x^i)]]_{(M \times M)} & y_{(M \times 1)} & [0]_{(M \times M)} & -[1]_{(M \times 1)} \\
-[\Upsilon [\phi(x^i)^T \phi(x^i)]]_{(M \times M)} & -y_{(M \times 1)} & -[I]_{(M \times M)} & [0]_{(M \times 1)}
\end{pmatrix} \label{eqn:kmcm_nn_3} \\
p=\begin{pmatrix}
[0]_{(M \times 1)} & -[1]_{(M \times 1)}
\end{pmatrix}^T \label{eqn:kmcm_nn_4}
\end{gather}
\end{figure*}

Also, in terms of kernel matrix $K(x^i,x^j)=K((x^i)^T, x^j)=[\phi(x^i)^T \phi(x^j)]$, the matrix $G$ can also be written as shown in Equation (\ref{eqn:kmcm_nn_5}).
\begin{figure*}
\begin{equation}
G=\begin{pmatrix}
[\Upsilon \cdot K(x^i,x^j)]_{(M \times M)} & y_{(M \times 1)} & [0]_{(M \times M)} & -[1]_{(M \times 1)} \\
-[\Upsilon \cdot K(x^i,x^j)]_{(M \times M)} & -y_{(M \times 1)} & -[I]_{(M \times M)} & [0]_{(M \times 1)}
\end{pmatrix} \label{eqn:kmcm_nn_5}
\end{equation}
\end{figure*}

These can be substituted in Equations (\ref{mcmdf01})-(\ref{mcmdf02}) and the system can be solved to obtain the equilibrium solution for the kernel case.

\section{Simulations of the MCM Neurodynamical System \label{sec4}}

\begin{figure*}[hbtp]
\centering
\begin{subfigure}[b]{0.49\textwidth}
 \includegraphics[width=\textwidth]{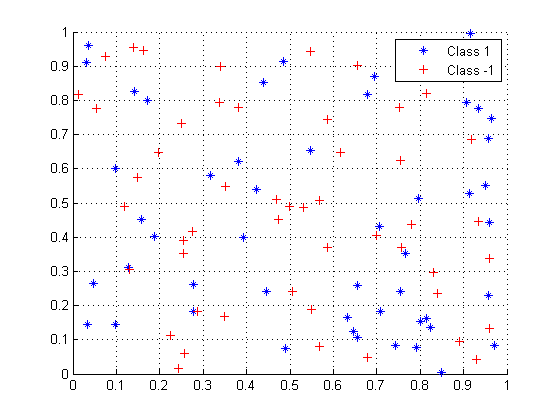}
 \caption{A Linearly Separable Dataset}
 \label{fig:dataset2}
\end{subfigure}
\begin{subfigure}[b]{0.49\textwidth}
 \includegraphics[width=\textwidth]{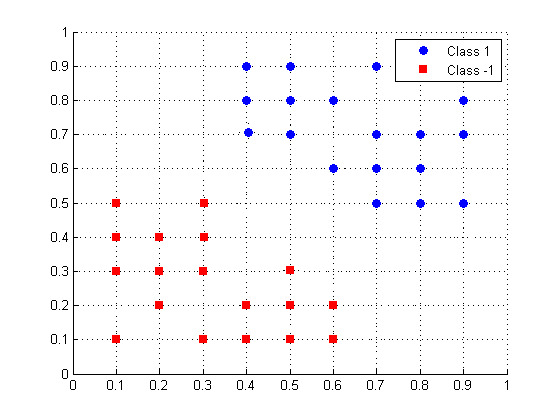}
 \caption{Dataset with points drawn from normal distribution}
 \label{fig:dataset1}
\end{subfigure}
\caption{(a) Example of a linearly separable dataset, and (b) a dataset with points drawn from a normal distribution.}
\label{fig:datasets}
\end{figure*}

In order to visualize the convergence of the system of differential equations, we provide the plots showing the evolution of the decision variables of our system, namely $w_i$'s, $b$ and $h$ over time. We consider the case for two datasets (both two dimensional), a linearly separable dataset shown in Fig. \ref{fig:dataset2} and a dataset with points (belonging to the two classes) randomly drawn from a normal distribution, as shown in Fig. \ref{fig:dataset1}.

\begin{figure*}
\centering
\begin{subfigure}[b]{0.49\textwidth}
 \includegraphics[width=\textwidth]{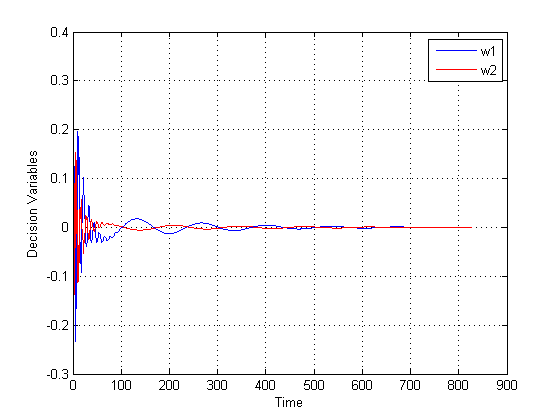}
 \caption{Plot of $w_1$ and $w_2$}
 \label{fig:dataset1_w1}
\end{subfigure}
\begin{subfigure}[b]{0.49\textwidth}
 \includegraphics[width=\textwidth]{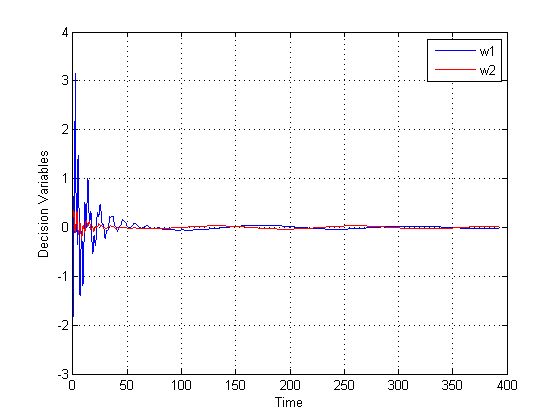}
 \caption{Plot of $\dot{w_1}$ and $\dot{w_2}$}
 \label{fig:dataset1_w2}
\end{subfigure}
\begin{subfigure}[b]{0.49\textwidth}
 \includegraphics[width=\textwidth]{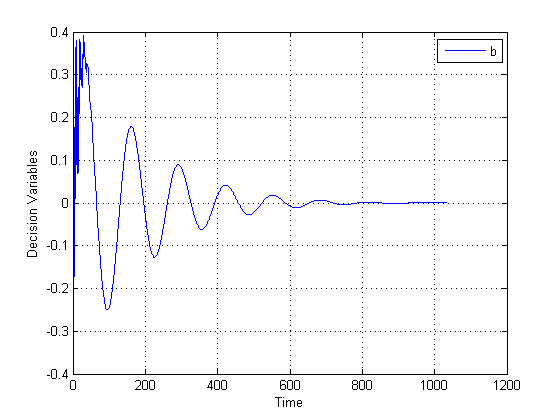}
 \caption{Plot of $b$}
 \label{fig:dataset1_b1}
\end{subfigure}
\begin{subfigure}[b]{0.49\textwidth}
 \includegraphics[width=\textwidth]{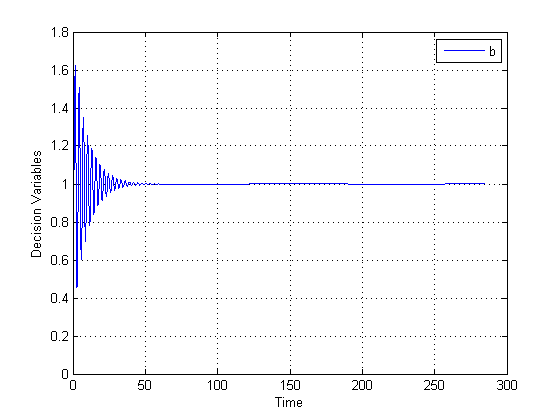}
 \caption{Plot of $\dot{b}$}
 \label{fig:dataset1_b2}
\end{subfigure}
\begin{subfigure}[b]{0.48\textwidth}
 \includegraphics[width=\textwidth]{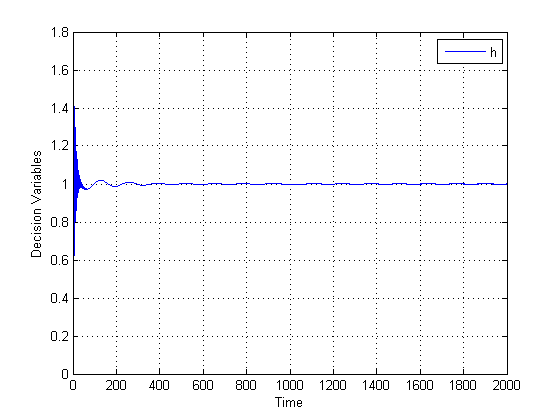}
 \caption{Plot of $h$}
 \label{fig:dataset1_h1}
\end{subfigure}
\begin{subfigure}[b]{0.48\textwidth}
 \includegraphics[width=\textwidth]{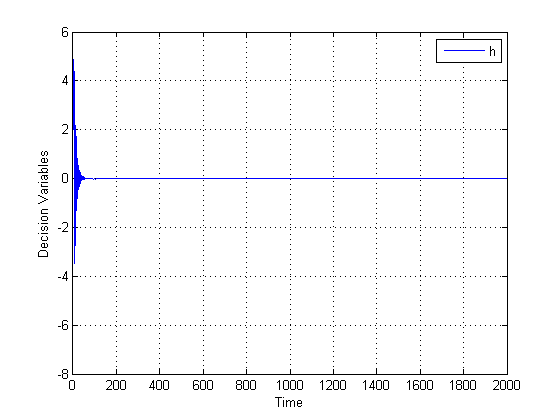}
 \caption{Plot of $\dot{h}$}
 \label{fig:dataset1_h2}
\end{subfigure}
\caption{Plots of convergence of the decision variables $w$, $b$ and $h$, and their first derivatives $\dot{w}$, $\dot{b}$ and $\dot{h}$ with time, for the linearly separable dataset shown in Fig. \ref{fig:dataset2}.}
\label{fig:dataset1_plots}
\end{figure*}

The plots for the decision variables for the linearly separable dataset are shown in Fig. \ref{fig:dataset1_plots}. The horizontal axis indicates the time in milliseconds. Figs. \ref{fig:dataset1_w1} and \ref{fig:dataset1_w2} show the plots of $w_1$, $w_2$ and their derivatives $\dot{w_1}$, $\dot{w_2}$ respectively. Plots of convergence of $b$ and its derivative $\dot{b}$ are shown in Figs. \ref{fig:dataset1_b1} and \ref{fig:dataset1_b2}, whereas those for $h$ and its derivative $\dot{h}$ are shown in Figs. \ref{fig:dataset1_h1} and \ref{fig:dataset1_h2}.

\begin{figure*}
\centering
\begin{subfigure}[b]{0.49\textwidth}
 \includegraphics[width=\textwidth]{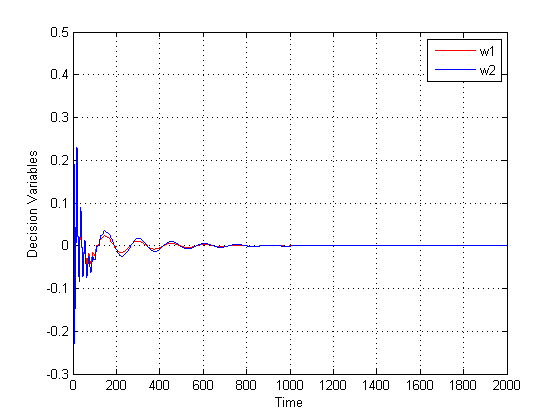}
 \caption{Plot of $w_1$ and $w_2$}
 \label{fig:dataset2_w1}
\end{subfigure}
\begin{subfigure}[b]{0.49\textwidth}
 \includegraphics[width=\textwidth]{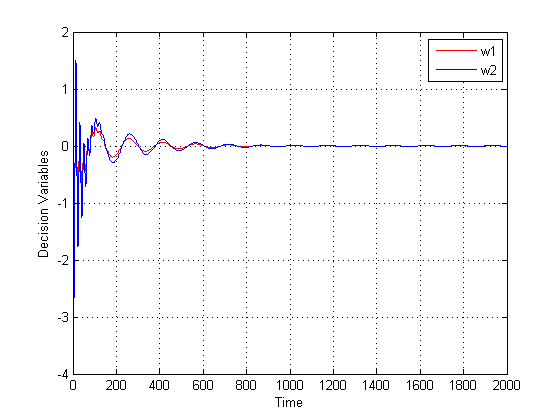}
 \caption{Plot of $\dot{w_1}$ and $\dot{w_2}$}
 \label{fig:dataset2_w2}
\end{subfigure}
\begin{subfigure}[b]{0.49\textwidth}
 \includegraphics[width=\textwidth]{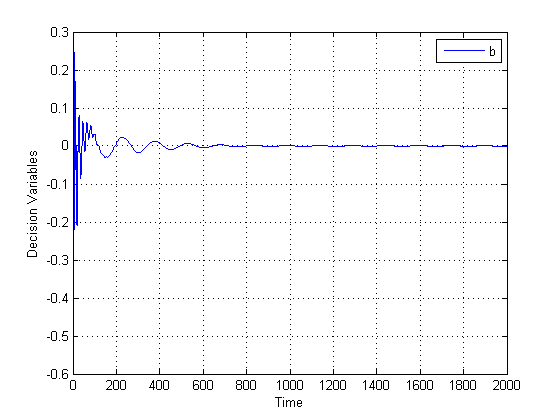}
 \caption{Plot of $b$}
 \label{fig:dataset2_b1}
\end{subfigure}
\begin{subfigure}[b]{0.49\textwidth}
 \includegraphics[width=\textwidth]{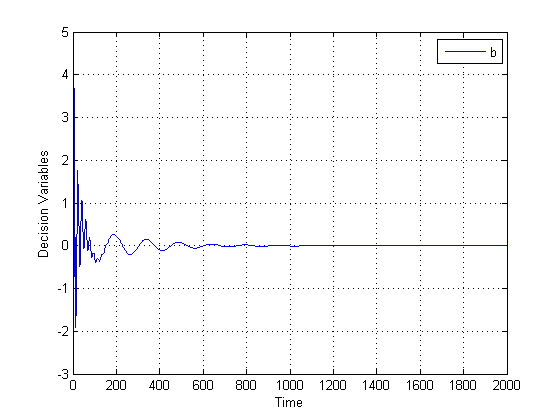}
 \caption{Plot of $\dot{b}$}
 \label{fig:dataset2_b2}
\end{subfigure}
\begin{subfigure}[b]{0.49\textwidth}
 \includegraphics[width=\textwidth]{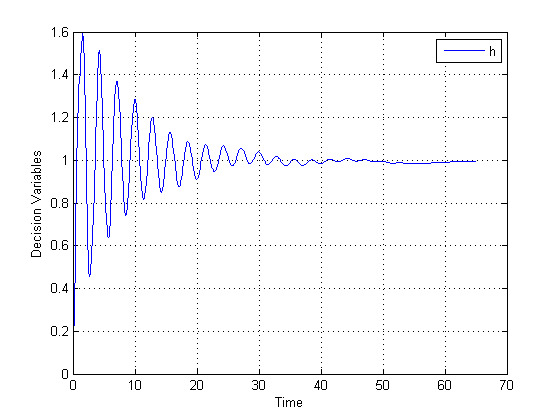}
 \caption{Plot of $h$}
 \label{fig:dataset2_h1}
\end{subfigure}
\begin{subfigure}[b]{0.49\textwidth}
 \includegraphics[width=\textwidth]{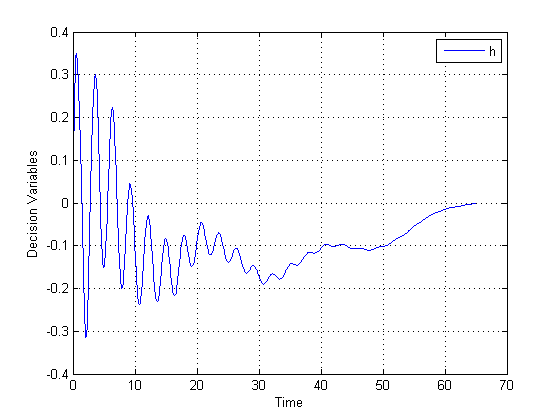}
 \caption{Plot of $\dot{h}$}
 \label{fig:dataset2_h2}
\end{subfigure}
\caption{Plots of convergence of the decision variables $w$, $b$ and $h$, and their first derivatives $\dot{w}$, $\dot{b}$ and $\dot{h}$ with time, for the dataset with points drawn from a normal distribution, as shown in Fig. \ref{fig:dataset1}}
\label{fig:dataset2_plots}
\end{figure*}

The plots for the decision variables for the dataset with points drawn from a normal distribution are shown in Fig. \ref{fig:dataset2_plots}. The horizontal axis represents time in milliseconds. Figs. \ref{fig:dataset2_w1} and \ref{fig:dataset2_w2} show the plots of $w_1$, $w_2$ and their derivatives $\dot{w_1}$, $\dot{w_2}$ respectively. Plots of convergence of $b$ and its derivative $\dot{b}$ are shown in Figs. \ref{fig:dataset2_b1} and \ref{fig:dataset2_b2}, whereas those for $h$ and its derivative $\dot{h}$ are shown in Figs. \ref{fig:dataset2_h1} and \ref{fig:dataset2_h2}.


\section{Results}\label{results}

The MCM neurodynamical system was implemented in Matlab vR2013a and the code executed on a laptop running 64-bit Windows Operating System with Intel i3 processors @2.53 Ghz and 4 Gb RAM. 

Table \ref{result_linear} shows the performance of the linear MCM dynamical system on a set of benchmark datasets from the UCI machine learning repository. The table also provides comparison with the standard SVM formulation in the linear case. For results in case of the linear MCM, see \cite{ jd2014b}. Accuracies are shown in a mean $\pm$ standard deviation format, computed using a standard five fold cross validation methodology. One can see that the MCM dynamical system outperforms the standard SVM in terms of test set accuracies.

\begin{table*}[!htbp]
  \centering
  \caption{Test Set Accuracies for the Linear MCM Dynamical System} 
  \scalebox{0.8}{
    \begin{tabular}{|c|c|c|c|c|}
    \hline
    S. No. & Dataset & Size (samples $\times$ features) & Linear MCM Dynamical System &  Linear SVM \\
    \hline
     1 & Hayes Roth &  132 $\times$ 5 & \textbf{76.11} $\pm$ 8.72 & 73.56 $\pm$ 7.73 \\
    2 & Hepatitis & 165 $\times$  19 & \textbf{69.35} $\pm$ 8.71 & 60.64 $\pm$ 7.19 \\
    3 & TA Evaluation & 151 $\times$ 5 & \textbf{69.52} $\pm$ 6.92 & 64.94 $\pm$ 6.56 \\
    4 & Promoters & 106 $\times$ 58 & \textbf{68.92 $\pm$ 6.91} &  67.78 $\pm$ 10.97 \\
    5 &  Voting & 435 $\times$ 16 & \textbf{95.97} $\pm$ 3.75 &  94.48 $\pm$ 2.46 \\
    6 & Australian & 690 $\times$ 14 & \textbf{85.79} $\pm$ 2.59 &  84.49 $\pm$ 1.18 \\
    7 & Bands & 512 $\times$ 39 & \textbf{72.58} $\pm$ 3.98 & 71.69 $\pm$ 3.81 \\
    8 & Fertility & 100 $\times$ 10 & 86.00 $\pm$ \textbf{6.91} & 86.00 $\pm$ 9.01 \\
    9 & Spect & 267 $\times$ 22 & 91.46 $\pm$ \textbf{4.28} &  91.99 $\pm$ 4.90 \\
    10 & Haberman & 306 $\times$ 3 & 72.01 $\pm$ \textbf{3.54} &  72.56 $\pm$ 3.73 \\
    11 & Planning-Relax & 182 $\times$ 13 & \textbf{72.41} $\pm$ 7.81 & 71.42 $\pm$ 7.37 \\
    \hline
    \end{tabular}%
    }
  \label{result_linear}%
\end{table*}%
Table \ref{result_kernel} shows the performance of the kernel MCM dynamical system on a set of benchmark datasets from the UCI machine learning repository. The table also provides a comparison with SVM using the RBF kernel. The hyper-parameter $C$ was tuned by using grid search. A similar search was used to determine the width of the RBF kernel. As indicated previously, accuracies are shown in mean $\pm$ standard deviation format, computed using a standard five fold cross validation methodology. One can see that the MCM dynamical system yields comparable or better performance than the SVM. Further, it is observed that the kernel MCM always uses fewer support vectors; indeed, up to \textbf{74.3\%} fewer support vectors (computed on the average number of support vectors). It may be noted that the number of support vectors presented in Table \ref{result_kernel} have been shown in the mean $\pm$ standard deviation format, across the folds on which the accuracies have been computed, and hence the values are shown as floating point numbers. Observe that in rows \textbf{1, 2, 4, 6, 7, 8, 9, 10} of Table \ref{result_kernel}, the proposed kernel MCM achieves a higher test set accuracy with a smaller number of support vectors than the standard kernel SVM. Since the number of support vectors has a significant bearing on the number of computations, the MCM can be seen to be parsimonious in terms of computational requirements. This also translates into lower power consumption figures in hardware and VLSI realizations \cite{chakrabartty2005sub, genov2003kerneltron}.


\begin{table*}[!htbp]
  \centering
  \caption{Test Set Accuracies and number of Support Vectors (\#SVs) for the Kernel MCM Dynamical System (KMCM-DS) compared with standard RBF Kernel SVM (KSVM)}
  \scalebox{0.75}{
    \begin{tabular}{|c|c|c||c|c||c|c|}
    \hline
    S. No. & Dataset & \pbox{20cm} {Size (samples $\times$ \\ features)} & \pbox{20cm}{ KMCM-DS \\ Test Set Acc.} & KMCM-DS \#SVs & \pbox{20cm}{KSVM \\ Test Set Acc.} & KSVM \#SVs\\
    \hline
    \hline
     1 & Spect & 267 $\times$ 22 & \textbf{91.99} $\pm$ 4.90 & \textbf{49.6 $\pm$ 0.54} & 84.21 $\pm$ 4.90 & 50.2 $\pm$ 9.88\\
     2 & TA Evaluation & 151 $\times$ 5 & \textbf{80.86} $\pm$ 6.87 & \textbf{26.60} $\pm$ 32.43 &  68.88 $\pm$ 6.48 & 86.00 $\pm$ 3.22\\
     3 & Fertility Diagnosis & 100 $\times$ 10 & 88.00 $\pm$ \textbf{1.03} & \textbf{9.80} $\pm$ 19.60 &  88.00 $\pm$ 9.27 & 38.20 $\pm$ 1.60 \\
     4 & Hayes Roth &132 $\times$ 5 & \textbf{81.45} $\pm$ 7.98 & \textbf{33.23 $\pm$ 1.11} &  79.57 $\pm$ 6.60 & 84.20 $\pm$ 2.04\\
     5 & Hepatitis &165 $\times$  19 & 79.35 $\pm$ \textbf{4.09}  & \textbf{20.00 $\pm$ 0.00} & 82.57 $\pm$ 6.32 & 72.20 $\pm$ 4.31 \\
     6 & Promoters & 106 $\times$ 58 & \textbf{69.87} $\pm$ 7.85 & \textbf{84.8 $\pm$ 0.44} & 66.45 $\pm$ 6.52 & 94.0 $\pm$ 0.70 \\
     7 & Bands & 512 $\times$ 39 &  \textbf{77.88} $\pm$ 4.14  & \textbf{341.2 $\pm$ 0.44} & 75.69 $\pm$ 3.81 & 427.6 $\pm$ 3.78\\
     8 & Planning-Relax & 182 $\times$ 13 & \textbf{78.57 $\pm$ 8.23} & \textbf{116.8 $\pm$ 0.54} & 71.42 $\pm$ 8.43 & 145.6 $\pm$ 6.45 \\
     9 & Haberman & 306 $\times$ 3 & \textbf{76.45 $\pm$ 4.37} & \textbf{71.0 $\pm$ 0.414} & 72.89 $\pm$ 4.58 & 137.4 $\pm$ 3.36 \\
     10 & Australian & 690 $\times$ 14 & \textbf{76.95} $\pm$ 2.63 & \textbf{152} $\pm$ 4.86 & 66.23 $\pm$ 1.84 & 244.8 $\pm$ 4.604 \\
    \hline
    \end{tabular}%
    }
  \label{result_kernel}%
\end{table*}%

\section{Conclusion}\label{conclusion}
  In this paper, we describe a Neurodynamical System that converges to a classifier with minimum VC dimension. A learning machine with such properties is attractive for building circuits that can exploit the advantages of speed and parallelism that neurodynamical systems offer. It is also of interest as part of larger learning networks and adaptive control systems. Further work in this direction involves developing neurodynamical systems using MCMs for regression and other classification scenarios such as multilabel and multiclass problems.

\bibliographystyle{plain}
\bibliography{mcm_linear_nn}

\begin{thebibliography}{10}

\bibitem{bernstein1995lyapunov}
Dennis~S Bernstein and Sanjay~P Bhat.
\newblock Lyapunov stability, semistability, and asymptotic stability of matrix
  second-order systems.
\newblock {\em Journal of Mechanical Design}, 117(B):145--153, 1995.

\bibitem{bhaya2006control}
Amit Bhaya and Eugenius Kaszkurewicz.
\newblock {\em Control perspectives on numerical algorithms and matrix
  problems}, volume~10.
\newblock SIAM, 2006.

\bibitem{brockett1988dynamical}
Roger~W Brockett.
\newblock Dynamical systems that sort lists, diagonalize matrices and solve
  linear programming problems.
\newblock In {\em Decision and Control, 1988., Proceedings of the 27th IEEE
  Conference on}, pages 799--803. IEEE, 1988.

\bibitem{burges1998}
Christopher~JC Burges.
\newblock A tutorial on support vector machines for pattern recognition.
\newblock {\em Data mining and knowledge discovery}, 2(2):121--167, 1998.

\bibitem{chakrabartty2005sub}
Shantanu Chakrabartty and Gert Cauwenberghs.
\newblock Sub-microwatt analog vlsi support vector machine for pattern
  classification and sequence estimation.
\newblock In {\em Advances in Neural Information Processing Systems 17:
  Proceedings of the 2004 Conference}, volume~17, page 249. MIT Press, 2005.

\bibitem{chua1984nonlinear}
Leon~O Chua and Gui-Nian Lin.
\newblock Nonlinear programming without computation.
\newblock {\em Circuits and Systems, IEEE Transactions on}, 31(2):182--188,
  1984.

\bibitem{chukwunenyeinterior}
Ukwu Chukwunenye.
\newblock On interior-point methods, related dynamical systems results and
  cores of targets for linear programming.
\newblock {\em International Journal of Mathematics and Statistics Invention},
  2(4):12--22, 2014.

\bibitem{corchs2001neurodynamical}
Silvia Corchs and Gustavo Deco.
\newblock A neurodynamical model for selective visual attention using
  oscillators.
\newblock {\em Neural Networks}, 14(8):981--990, 2001.

\bibitem{L1svm}
Corinna Cortes and Vladimir Vapnik.
\newblock Support-vector networks.
\newblock {\em Machine learning}, 20(3):273--297, 1995.

\bibitem{dobosz2010understanding}
Krzysztof Dobosz and W{\l}odzis{\l}aw Duch.
\newblock Understanding neurodynamical systems via fuzzy symbolic dynamics.
\newblock {\em Neural Networks}, 23(4):487--496, 2010.

\bibitem{faybusovich1991dynamical}
L~Faybusovich.
\newblock Dynamical systems which solve optimization problems with linear
  constraints.
\newblock {\em IMA Journal of Mathematical Control and Information},
  8(2):135--149, 1991.

\bibitem{faybusovich1992dynamical}
L~Faybusovich.
\newblock Dynamical systems that solve linear programming problems.
\newblock In {\em Decision and Control, 1992., Proceedings of the 31st IEEE
  Conference on}, pages 1626--1631. IEEE, 1992.

\bibitem{faybusovich1991hamiltonian}
Leonid Faybusovich.
\newblock Hamiltonian structure of dynamical systems which solve linear
  programming problems.
\newblock {\em Physica D: Nonlinear Phenomena}, 53(2):217--232, 1991.

\bibitem{freeman2007indirect}
Walter~J Freeman.
\newblock Indirect biological measures of consciousness from field studies of
  brains as dynamical systems.
\newblock {\em Neural Networks}, 20(9):1021--1031, 2007.

\bibitem{genov2003kerneltron}
Roman Genov and Gert Cauwenberghs.
\newblock Kerneltron: support vector" machine" in silicon.
\newblock {\em Neural Networks, IEEE Transactions on}, 14(5):1426--1434, 2003.

\bibitem{ghasabi2006efficient}
Hasan Ghasabi-Oskoei and Nezam Mahdavi-Amiri.
\newblock An efficient simplified neural network for solving linear and
  quadratic programming problems.
\newblock {\em Applied Mathematics and Computation}, 175(1):452--464, 2006.

\bibitem{hasegawa2002solving}
Mikio Hasegawa, Tohru Ikeguchi, and Kazuyuki Aihara.
\newblock Solving large scale traveling salesman problems by chaotic
  neurodynamics.
\newblock {\em Neural Networks}, 15(2):271--283, 2002.

\bibitem{hu2012solving}
Xiaolin Hu and Jun Wang.
\newblock Solving the assignment problem using continuous-time and
  discrete-time improved dual networks.
\newblock {\em Neural Networks and Learning Systems, IEEE Transactions on},
  23(5):821--827, 2012.

\bibitem{jd2014b}
Jayadeva.
\newblock Learning a hyperplane classifier by minimizing an exact bound on the
  \{VC\} dimension.
\newblock {\em Neurocomputing}, 149, Part B(0):683 -- 689, 2015.

\bibitem{mcm_featsel}
Jayadeva, Sanjit~S. Batra, and Siddharth Sabharwal.
\newblock Feature selection through minimization of the vc dimension.
\newblock {\em CoRR}, abs/1410.7372, 2014.

\bibitem{mcm_fuzzy}
Jayadeva, Sanjit~S. Batra, and Siddharth Sabharwal.
\newblock Feature selection through minimization of the vc dimension.
\newblock {\em CoRR}, abs/1501.02432, 2015.

\bibitem{mcm_regress}
Jayadeva, Suresh Chandra, Sanjit~S. Batra, and Siddharth Sabharwal.
\newblock Learning a hyperplane regressor by minimizing an exact bound on the
  vc dimension.
\newblock {\em CoRR}, abs/1410.4573, 2014.

\bibitem{kennedy1988neural}
Michael~Peter Kennedy and Leon~O Chua.
\newblock Neural networks for nonlinear programming.
\newblock {\em Circuits and Systems, IEEE Transactions on}, 35(5):554--562,
  1988.

\bibitem{liu2013one}
Qingshan Liu and Jun Wang.
\newblock A one-layer projection neural network for nonsmooth optimization
  subject to linear equalities and bound constraints.
\newblock {\em Neural Networks and Learning Systems, IEEE Transactions on},
  24(5):812--824, 2013.

\bibitem{maa1992linear}
C-Y Maa and Michael~A Shanblatt.
\newblock Linear and quadratic programming neural network analysis.
\newblock {\em Neural Networks, IEEE Transactions on}, 3(4):580--594, 1992.

\bibitem{mangasarian1992neural}
K~Mangasarian.
\newblock Neural network training via linear programming.
\newblock {\em Advances in Optimisation and Parallel Computing}, pages 56--67,
  1992.

\bibitem{marupaka2012connectivity}
Nagendra Marupaka, Laxmi~R Iyer, and Ali~A Minai.
\newblock Connectivity and thought: The influence of semantic network structure
  in a neurodynamical model of thinking.
\newblock {\em Neural Networks}, 32:147--158, 2012.

\bibitem{nguyan2000nonlinear}
KV~Nguyen.
\newblock A nonlinear neural network for solving linear programming problems.
\newblock In {\em ISMP-2000, 17-th International Symposium on Mathematical
  Programming}, pages 7--11.

\bibitem{pascanu2011neurodynamical}
Razvan Pascanu and Herbert Jaeger.
\newblock A neurodynamical model for working memory.
\newblock {\em Neural networks}, 24(2):199--207, 2011.

\bibitem{perez2013new}
Mar{\'\i}a~Jos{\'e} P{\'e}rez-Ilzarbe.
\newblock New discrete-time recurrent neural network proposal for quadratic
  optimization with general linear constraints.
\newblock {\em IEEE transactions on neural networks and learning systems},
  24(2):322--328, 2013.

\bibitem{rodriguez1990nonlinear}
Angel Rodriguez-Vazquez, Rafael Dominguez-Castro, Adoraci{\'o}n Rueda, Jose~L
  Huertas, and Edgar Sanchez-Sinencio.
\newblock Nonlinear switched capacitorneural'networks for optimization
  problems.
\newblock {\em Circuits and Systems, IEEE Transactions on}, 37(3):384--398,
  1990.

\bibitem{suykens1999least}
Johan~AK Suykens and Joos Vandewalle.
\newblock Least squares support vector machine classifiers.
\newblock {\em Neural processing letters}, 9(3):293--300, 1999.

\bibitem{tank1986simple}
Df~Tank and John~J Hopfield.
\newblock Simple'neural'optimization networks: An a/d converter, signal
  decision circuit, and a linear programming circuit.
\newblock {\em Circuits and Systems, IEEE Transactions on}, 33(5):533--541,
  1986.

\bibitem{vapnik98}
Vladimir Vapnik.
\newblock {\em Statistical learning theory}.
\newblock Wiley, 1998.

\bibitem{vapnik1974theory}
Vladimir~N Vapnik and A~Ja Chervonenkis.
\newblock Theory of pattern recognition.
\newblock 1974.

\bibitem{wang1993analysis}
Jun Wang.
\newblock Analysis and design of a recurrent neural network for linear
  programming.
\newblock {\em Circuits and Systems I: Fundamental Theory and Applications,
  IEEE Transactions on}, 40(9):613--618, 1993.

\bibitem{wu1996high}
Xin-Yu Wu, You-Shen Xia, Jianmin Li, and Wai-Kai Chen.
\newblock A high-performance neural network for solving linear and quadratic
  programming problems.
\newblock {\em Neural Networks, IEEE Transactions on}, 7(3):643--651, 1996.

\bibitem{xia1996new}
Youshen Xia.
\newblock A new neural network for solving linear and quadratic programming
  problems.
\newblock {\em Neural Networks, IEEE Transactions on}, 7(6):1544--1548, 1996.

\bibitem{xia2012discrete}
Youshen Xia, Changyin Sun, and Wei~Xing Zheng.
\newblock Discrete-time neural network for fast solving large linear estimation
  problems and its application to image restoration.
\newblock {\em Neural Networks and Learning Systems, IEEE Transactions on},
  23(5):812--820, 2012.

\bibitem{xia1995neural}
Youshen Xia and Jiasong Wang.
\newblock Neural network for solving linear programming problems with bounded
  variables.
\newblock {\em Neural Networks, IEEE Transactions on}, 6(2):515--519, 1995.

\bibitem{xia1998general}
Youshen Xia and Jun Wang.
\newblock A general methodology for designing globally convergent optimization
  neural networks.
\newblock {\em Neural Networks, IEEE Transactions on}, 9(6):1331--1343, 1998.

\bibitem{yan2014collective}
Zheng Yan, Jun Wang, and Guocheng Li.
\newblock A collective neurodynamic optimization approach to bound-constrained
  nonconvex optimization.
\newblock {\em Neural Networks}, 55:20--29, 2014.

\end{thebibliography}

\end{document}